\documentclass[preprint]{article}

\usepackage[nonatbib]{neurips_2024}

\usepackage[utf8]{inputenc} 
\usepackage[T1]{fontenc}    
\usepackage{hyperref}       
\usepackage{url}            
\usepackage{booktabs}       
\usepackage{nicefrac}       
\usepackage{microtype}      
\usepackage{xcolor}         

\usepackage{amsmath,amsfonts}
\usepackage{algorithmic}
\usepackage{algorithm}
\usepackage{array}
\usepackage[caption=false,font=normalsize,labelfont=sf,textfont=sf]{subfig}
\usepackage{textcomp}
\usepackage{stfloats}
\usepackage{verbatim}
\usepackage{graphicx}
\usepackage{cite}
\usepackage{bbding}

\title{Learning Structurally Stabilized Representations for Multi-modal Lossless DNA Storage}

\author{%
  Ben Cao\\
  Dalian University of Technology\\
  School of Computer Science and Technology\\
  A*STAR, CFAR \\
  \texttt{Bencaocs@gmail.com} \\
  \And
  Tiantian He $^\dagger$\\
  A*STAR\\
  CFAR\&SIMTECH \\
  \texttt{He\_Tiantian@cfar.a-star.edu.sg} \\
  \And
  Xue Li \\
  UPC \& Duke-NUS \\
  \texttt{Xueleecs@gmail.com} \\
  \And
  Bin Wang \\
  Dalian University \\
  School of Software Engineering\\
  \texttt{Wangbinpaper@gmail.com} \\
  \And
  Xiaohu Wu\\
  BUPT\\
  \texttt{Xiaohu.wu@bupt.edu.cn}
  \And
  Qiang Zhang $^\dagger$ \\
  Dalian University of Technology\\
  School of Computer Science and Technology\\
  \texttt{Zhangq@dlut.edu.cn}\\
  \And
  Yew-Soon Ong\\
  Nanyang Technology University\\
   School of Computer Science and Engineering\\
  \texttt{ASYSOng@ntu.edu.sg} \\
}

\begin{document}
\setlength{\abovecaptionskip}{-5pt}
\setlength{\belowcaptionskip}{-5pt}

\maketitle

\begin{abstract}
In this paper, we present Reed-Solomon coded single-stranded representation learning (RSRL), a novel end-to-end model for learning representations for  multi-modal lossless DNA storage.
In contrast to existing learning-based methods, the proposed RSRL is inspired by both error-correction codec and structural biology.
Specifically, RSRL first learns the representations for the subsequent storage from the binary data transformed by the Reed-Solomon codec.
Then, the representations are masked by an RS-code-informed mask to focus on correcting the burst errors occurring in the learning process.
With the decoded representations with error corrections, a novel biologically stabilized loss is formulated to regularize the data representations to possess stable single-stranded structures.
By incorporating these novel strategies, the proposed RSRL can learn highly durable, dense, and lossless representations for the subsequent storage tasks into DNA sequences.
The proposed RSRL has been compared with a number of strong baselines in real-world tasks of multi-modal data storage.
The experimental results obtained demonstrate that RSRL can store diverse types of data with much higher information density and durability but much lower error rates.

\end{abstract}

\renewcommand{\thefootnote}{\fnsymbol{footnote}}
\footnotetext[2]{Corresponding author.}

\section{Introduction}
DNA storage has become one of the most promising technical solutions for coping with data explosion \cite{Church2012,Erlich2017,ping2022towards}.
Compared with conventional storage techniques, DNA storage utilizes DNA molecules as a storage medium to read and write data.
By integrating advanced bio-technologies, such as DNA coding, synthesis, sequencing, recovery, and decoding, it achieves desirable characteristics of high-density \cite{ChenNSR2021}, high-durability \cite{nguyen2021capacity}, and ultra-long-time storage \cite{wang2023modelling}.
Although benefiting from the emergence of modern information and biotechnology \cite{Carmean2019ProIEEE}, DNA storage still suffers from the critical bottlenecks of cost and latency compared with electromagnetic storage media.

Recently, leveraging computational approaches to break through the cost and latency bottlenecks of existing DNA storage has drawn much attention from AI and machine learning communities.
Several models for DNA coding and decoding have been developed to learn compact data representations that can improve the base utilization \cite{Zan2023JCIM} while reducing latency issues \cite{qu2022clover}.
These approaches can be categorized into two classes, i.e., coding-theory-based and learning-based approaches.
Methods based on coding theory \cite{Grass2015,Goldman2013,Anavy2019,nguyen2021capacity} are designed to strictly follow certain coding systems.
Thus, they have high storage capacity ratios.
However, these approaches are computationally demanding when dealing with large-scale data.
In contrast, learning-based approaches adopt heuristic searching algorithms \cite{Lochel2021nar} or deep neural networks \cite{li2022hldna,franzese2021generative,wu2023deepJoint} to acquire an optimized coder/decoder that can write or read data stored in the DNA sequences.
Though effective to some extent, these learning-based approaches always suffer from limited base utilization.
Besides, they lack sufficient biological constraints during training, which can compromise data integrity. Thus, they are applicable only to data types that can tolerate information loss, such as images and videos.

In this paper, we hypothesize that the coalescence of contemporary learning models and stable traits of biomolecular structures in DNA can overcome the previously mentioned challenges confronted by existing learning-based approaches for DNA storage.
To this end, we present Reed-Solomon coded single-stranded representation learning (RSRL), a novel model for learning representations for lossless DNA storage of multi-modal data.
To develop RSRL, we make the following two technical contributions. Firstly, inspired by the Reed-Solomon codec, we propose a novel data preprocessing and mask strategy for representation learning for DNA storage. Specifically, the representations are learned from binary data coded based on the Reed-Solomon codec. Then, the representations are masked by an RS-code-informed Mask to focus on correcting the burst errors occurring in the learning process.
Secondly, with the decoded data representations with error corrections, we propose a novel biologically stabilized loss that regularizes the data representations to possess stable single-stranded structures.
With the mentioned techniques incorporated into the training process, the data representations learned for the subsequent writing to DNA are highly durable, dense, and lossless.
In our experiments, the proposed RSRL has been compared with several strong baselines in real-world tasks of multi-modal data storage. The results demonstrate that RSRL achieves a notable reduction in learning complexity, with an 18\% increase in net information density and an 11\% improvement in thermodynamic performance. Additionally, our approach reduces coding and decoding delays by more than two orders of magnitude, representing a significant advancement in the field of DNA storage technology.
The main contributions of this paper are summarized as follows:
\begin{itemize}
	\item We have verified that the consideration of stable traits of biomolecular structures and error-correction codec can significantly improve the representation learning for DNA storage.
 	\item To the best of our knowledge, we initialize the first attempt to formulate the loss function that can guide the proposed RSRL to learn representations for the subsequent DNA storage tasks that possess stable single-stranded structures.
    Such formulated loss enables RSRL to learn representations for durable, dense, and lossless DNA storage.
	\item The proposed RSRL has been compared with strong baselines in real-world storage tasks of multi-modal data.
    Experimental results show that RSRL can overcome the shortcomings of existing learning-based approaches, indicating that it is an effective and promising method for real-world DNA storage tasks.
 \end{itemize}

\section{Related work}

\paragraph{Conventional DNA storage and coding} Due to intrinsically bearing life information as a natural storage medium, DNA has become the most competitive alternative to silicon-based storage \cite{Limbachiya202210year}.
DNA storage can be divided into three main phases, including data writing \cite{rasool2023bo,Grass2015,bogels2023dna}, preservation, and reading \cite{cao2024efficient, Organick2018Nbt,bogels2023dna}. 
Conventionally, utilizing DNA molecules for data storage primarily includes two methods, i.e., sequence base representation and structure representation.
Most approaches are designed to capture the distribution of bases to convert abiotic information into DNA sequences through DNA coding.
Efficient and robust coding schemes not only improve coding efficiency but also ensure data integrity. 
However, due to the uncontrollability of biomolecules \cite{zhang2023biomolecule} and the inherent errors during DNA synthesis and sequencing, the coding rate was still some distance from the theoretical upper limit. 
Existing approaches to DNA storage coding can be divided into two categories, i.e., coding theory-based and learning-based \cite{nguyen2021capacity}. 
Huffman coding is the earliest framework of coding theory that is used in DNA coding for storage purposes \cite{Goldman2013}.
It can completely avoid consecutive base repetitions.
Subsequently, the Galois field and DNA codon wheel are combined for coding to avoid consecutive bases greater than three.
Recently, Reed-Solomon coding \cite{blawat2016forward}, prefix-synchronized coding \cite{Yazdi2015Rewritable}, DNA Fountain \cite{Erlich2017}, the ying-yang code \cite{ping2022towards}, and Repeat-Accumulate coding \cite{wang2023modelling} have also been considered for empirical DNA storage systems.

\paragraph{Learning-based DNA storage} On the other hand, the main idea of learning-based methods is to compress data vectors using neural networks and then encode the compressed vectors for storage \cite{franzese2021generative,guo2021using}.
One of the representatives is DNA-QLC \cite{Zhengyan2024BMC}, which adopts over ten layers of CNNs to extract hidden information from images and encode it into DNA sequences using Levenshtein code.
Recently, biological constraints have been considered when building learning-based models for DNA storage.
For example, homopolymer and GC content have been used to build the loss functions \cite{wu2023deepJoint}, leveraging which a CNN-based encoder-decoder can learn image representations for DNA storage.
Though effective, these learning-based approaches are either computationally demanding or only applicable for storing data that are tolerable for some information loss (e.g., images).
Differing from both traditional and learning-based approaches, the proposed RSRL considers coding theory and biologically stabilized structures when learning data representations.
This novel strategy enables RSRL to learn highly durable, dense, and lossless representations for DNA storage.

\section{The proposed method}
In this section, we introduce RSRL, a novel end-to-end model for learning representations for multi-modal lossless DNA storage. 
Fig.~1 illustrates the brief architecture of RSRL.
RS codec is firstly used to process multi-modal data into a redundant binary data stream to correct errors during the learning process. 
This data stream is then fed into a Transformer network \cite{Vaswani2017Transformer} to learn representations.
Based on the RS codec, we design a Mask-MSE loss that can correct the burst errors widely existing in DNA sequences for storage.
We further design a hairpin loss to ensure that the DNA codewords carrying non-biological information (i.e., the learned data representations) possess the stable single-strand structures possessed by DNA in the biont.
It is noted that the one-stranded structure is considered in RSRL due to the complementary pairing structure of DNA double helices.
The low-dimensional data representations are transcoded into DNA sequences and then automatically paired to form double-helix structures for data storage.

\begin{figure}[!t]
	\centering
	\includegraphics[width=0.8\textwidth]{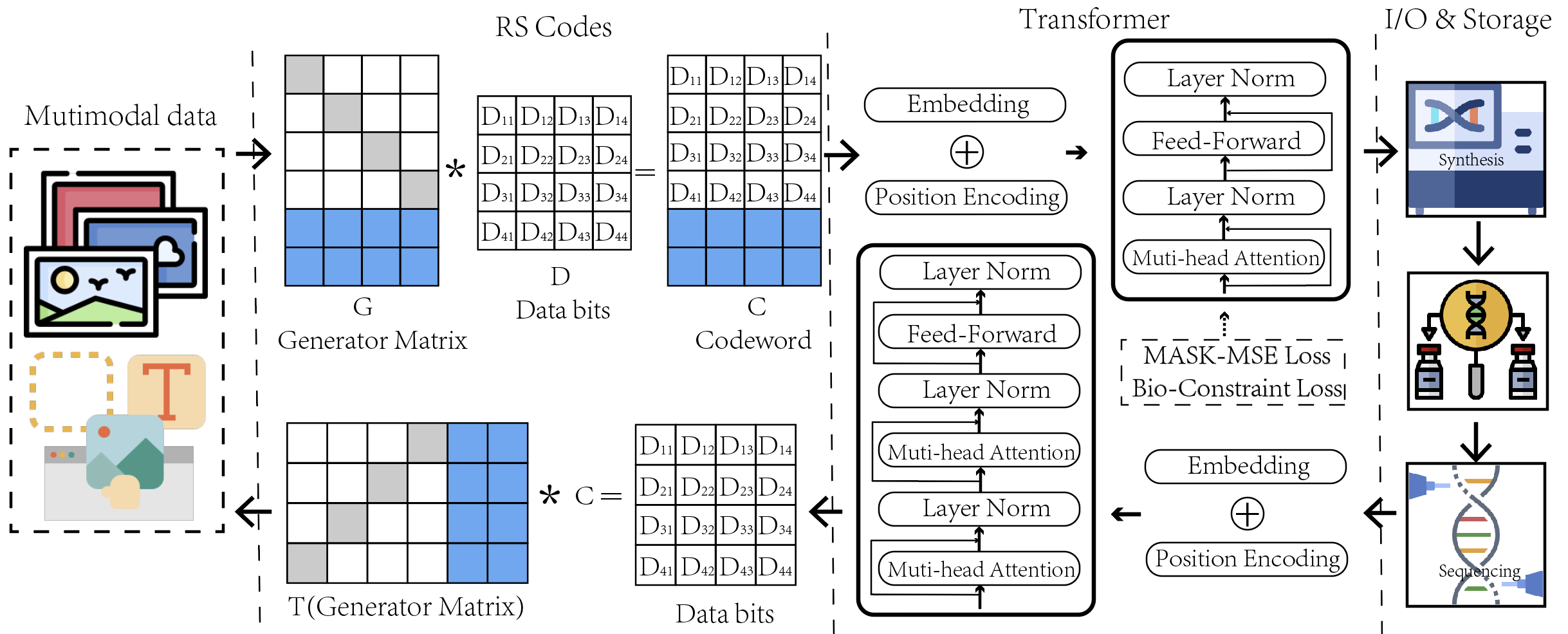}
        \vspace{1.0em}
	\caption{Overview of the DNA storage scheme implemented by RSRL.}
	\label{fig_1}
        
\end{figure}

\subsection{Representation learning and codec}

For any file of size, $W$ to be stored, it is first converted to a binary matrix $M*48$, $M=W/48$, and this matrix is $RS(64, 48)$ encoded by rows to obtain a binary data stream ($M*64$) with error-correction redundancy. After reshaping the data stream into $N*32*64$, $N=M/32$, a transformer with a compression layer \cite{Vaswani2017Transformer} is employed to extract low-dimensional data representations, which are encoded as DNA sequences for data storage after optimization by a loss function. 

Specifically, RS codes adopted in this paper are typically defined in Galois fields, given a finite field $F$ and polynomial ring $F[x]$, where $n$ and $k$ satisfy $1\leq k\leq n\leq|F|$. Selecting $n$ distinct elements from $F$, denote as $\{x_{1},x_{2},\ldots,x_{n}\}$. The codeword $\mathbf{C}$ is obtained by computing the values of polynomials in $F[x]$ such that the order of each $x_{i}$ in $F$ is less than $k$:
\begin{equation}
	\mathbf{C}=\left\{\left(f(x_1),f(x_2),\ldots,f(x_n)\right)|f\in F[x],\deg(f)<k\right\}.
\end{equation}
So $\mathbf{C}$ is an $[n,k,n-k+1]$ code, which is also a linear code in $F$ of length $n$, dimension $k$, and minimum Hamming distance $n-k+1$. 

Thus, any dimensionally matched binary matrix of the file to be stored can be RS encoded according to $\mathbf{C}$ to get a binary data stream with error-correction redundancy.
Then binary data stream is used as the input sequences $Y=\{y_{1},y_{2},\ldots,y_{n}\}$, where $y_{i}$ represents the data (representation) $i$ in some Transformer layer, the output representations that will be either fed into the loss functions or passed to next layers, are learned through the self-attention mechanism.
Specifically, three matrices queries ($Q = YW_q$), keys ($K = YW_k$), and values ($V = YW_v$) are firstly computed.
Based on them, the output representation for $Y$ can be generated as follows:
\begin{equation}
	\label{deqn_ex1a}
	\mathrm{Attention}(Q,K,V)=\mathrm{softmax}\left(\frac{QK^T}{\sqrt{d_k}}\right)V,
\end{equation}
where $d_{k}$ is the dimension of the query and key. 
After training with input data that have been encoded by the RS codec, RSRL is able to learn the compressed representations of data, which are then further encoded as sequences of nucleobase in DNA. 

DNA sequences (transcoded representation) should be biostable, which is required to integrably preserve the information carried by the low-dimensional representations.
To improve learning efficiency, complex encoding methods are not suitable for use during training. 
A widely accepted method for encoding representations to nucleobase is to directly map $00-A,01-T,11-G,10-C$. 
However, this coding method is prone to homopolymers.
Additional constraints on homopolymers are required by this coding method, thus increasing the complexity of the encoding.
In this paper, we propose a novel block mapping strategy for representations for DNA storage encoding. 
Specifically, we view two bases as a community and encode the four-bit representation into two bases at once (details can be checked in Table \ref{binary-code} in the Appendix).
This strategy can minimize the generation of homopolymers.
Results presented in Section \ref{experiment} validate its effectiveness.

\subsection{Biologically stabilized loss functions}
Existing loss functions fail to guide a learning model to achieve lossless DNA storage as they do not consider factors of biological stability.
Inspired by the single-stranded structure in RNA and RS codec, we propose to formulate biologically stabilized loss functions that can guide the learned representations to possess the stable structures like bio-molecules have, thus achieving highly durable, information-dense, and lossless storage in DNA.

\subsubsection{Synergizing RS codes with MASK-MSE loss}
The primary purpose of data storage is to ensure the consistency of data reading and writing. 
Naturally, mean squared error (MSE) is a widely accepted loss function that quantifies the average deviation between the reconstructed binary data stream and the original data. 
However, conventional MSE cannot fully address the errors caused by information loss during representation learning.
Moreover, to ensure data integrity, RSRL incorporates RS codes as an error-correction measure that is good at handling burst errors (a series of adjacent errors).
In contrast, random errors (random single-bit errors) generally exist in DNA storage channels.
Therefore, to fully utilize the error-correction capability of the RS codec, random errors in DNA storage have to be transformed into burst errors.
To address these two issues, we introduce an additional mask operation to MSE loss, and thus propose MASK-MSE loss based on RS codes, which guides the reduction of learning efficiency for a specific integer block of the current tensor during the learning process, concentrating errors within this block.

\begin{figure}[!t]
    \centering
    \begin{minipage}[t]{0.48\textwidth}
        \centering
        \includegraphics[width=\textwidth]{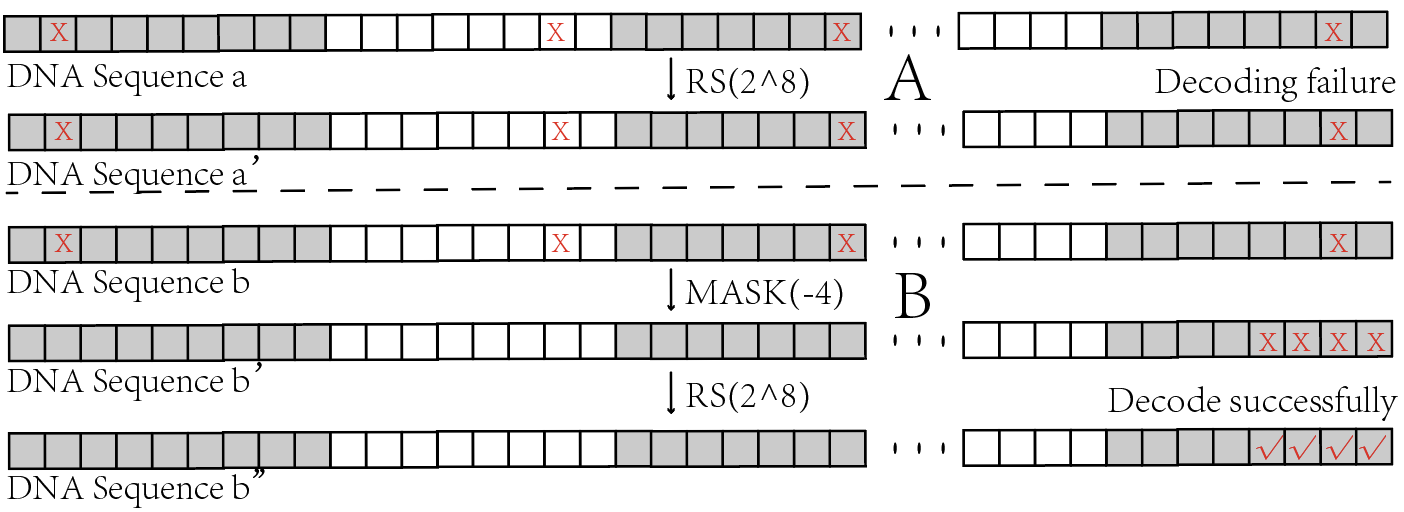}
        \caption{MASK-MSE loss Maximizes the potential of RS error correction codes.}
        \label{fig_2}
    \end{minipage}
    \hfill
    \begin{minipage}[t]{0.48\textwidth}
        \centering
        \includegraphics[width=\textwidth]{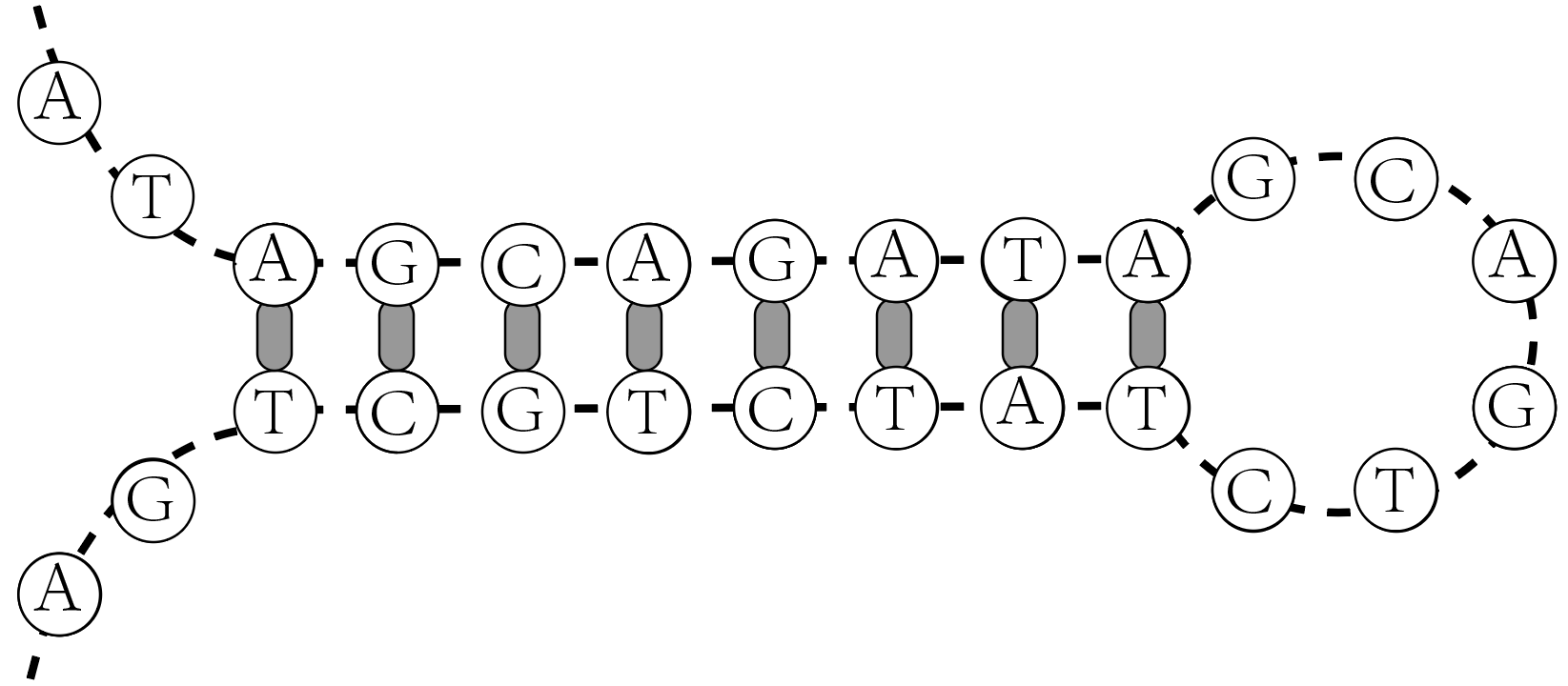}
        \caption{The Hair structure.}
        \label{fig_3}
    \end{minipage}
\end{figure}

Let $Z$ denote the ground truth data tensor, $\hat{Z}$ represents the predicted tensor (i.e., the output representation learned from $Y$), $\mathbf M$ and $N$ denote the mask tensor and the total number of elements in the tensor, $i$ and $j$ are the indices of the elements in the tensors.
$\mathbf M_{i,j}$ is the mask value at $position (i, j)$. If $\mathbf M_{i,j}=1$, the loss is computed at corresponding position. Otherwise, the loss regarding $Z_{i,j}$ is masked out:
\begin{equation}
	{\mathcal L}_{\mathrm{MASK-MSE}}=\frac{1}{N}\sum_{i,j}\mathbf M_{i,j}\cdot(Z_{i,j}-\hat{Z}_{i,j})^{2}.
\end{equation}
As backpropagation is deactivated when $Z_{i,j}$ is masked out, the RS codec will dominate the encoding process of $Z_{i,j}$, thus correcting potential errors during training.
In this paper, we set the mask size of each representation as eight, which is the same as the block size of the RS codes adopted by the proposed RSRL.
With the proposed MASK-MSE loss, dispersed random errors are transformed into burst errors, which RS is adept at handling (Fig. \ref{fig_2}).

\subsubsection{Learning single-stranded representations}
It is known that DNA and RNA with stable structures can generally carry genetic information with minimized errors in transcription and translation \cite{leppek2022combinatorial}.
In this paper, we introduce single-stranded loss functions to endow the learned representations with the previously mentioned properties of stable structures.
Here, we formulate a single-stranded loss by mainly considering GC content and hairpin structure. 
The proposed single-stranded loss will be computed based on the DNA sequences that are transcoded from the learned representations according to Table \ref{binary-code} in the Appendix (i.e., sequence $l$ is $\hat{Z}_l$ in $\hat{Z}$ after transcoding). 
Let $\mathcal{G}(l)$ and $\mathcal{H}(l)$ denote the GC content and hairpin structure of the sequence $l$,
$\mathcal{G}^*$ and $\mathcal{H}^*$ denote the target values of $\mathcal{G}(l)$ and $\mathcal H(l)$, which are 50\% and 0, respectively.
Therefore, our goal is to formulate a loss that can minimize the difference between $\mathcal G(l)$ and $\mathcal G^*$, and that between $\mathcal H(l)$ and $\mathcal H^*$.

Since there are more hydrogen bonds between bases G\&C than between A\&T and keeping the bases evenly distributed is beneficial to the stability of the DNA sequence \cite{ping2022towards}, we use the GC content as one of the learning objectives. The GC content of each DNA sequence transcoded from the corresponding representation is computed as the following: 
\begin{equation}\label{gc}
	\mathcal G(l)=\frac{|G|+|C|}{|G|+|C|+|A|+|T|}\times100\%,
\end{equation}
where $|\cdot|$ is defined as the sum of the number of bases in each DNA sequence. 

A hairpin structure forms a hairpin-like shape in which two base pairs are bonded together by hydrogen bonds to form a loop (Fig. \ref{fig_3}), which increases the error rate in reading and replicating DNA storage data \cite{li2020constraining}. 
Therefore, we aim to form a loss that can minimize the hairpin structure in learned representations.
Hairpin structures have two important parameters, i.e., the minimum stem region $S_{min}$ and the minimum ring region $R_{min}$. In calculating the probability of forming a hairpin structure of different sizes at each position of the sequence, the first consideration is the result of forming a hairpin at position $i$ with $R$ ring and $S$ stem region. A hairpin structure is considered to be formed if more than half of $l_{i-s}\cdots l_{i}$ and $l_{i+r}\cdots l_{i+r+s}$ are hybridized. 
For each sequence transcoded from the learned representation, the number of existing hairpin structures can be computed as follows: 
\begin{equation}\label{hairpin}
	\begin{split}
		\mathcal H(l) = & \sum_{s=S_{\mathrm{min}}}^{(L-R\mathrm{min})/2} \sum_{r=R_{\mathrm{min}}}^{L-2s}  \sum_{i=1}^{L-2s-r} T\left(\sum_{j=1}^{s} bp\left(l_{s+i-j},l_{s+i+r+j-1}\right),\frac{s}{2}\right),
	\end{split}
\end{equation}
where $s$ is the stem length, $S_{min}$is the set minimum stem length. $r$ is the ring length, $R_{min}$ is the set minimum ring length, and $L$ denotes the length of the DNA sequence. $T(\cdot)$ is the threshold function. Only when $m>n$, $T(m, n) = m$, otherwise = 0, the threshold is generally taken as 0.5, i.e., only when the consecutive matches reached the half of the stem region is regarded as the generation of the hairpin structure. $bp(j, k)$ function indicates the number of bases complementary to each other at positions $j$ and $k$ in the DNA sequence; complementary is 1. Otherwise, it is 0. The ranges of values of the relevant parameters are as follows, the maximum length of the stem region $(L-R_{min})/2$ is obtained at $R_{min}$, and the maximum length of the loop region $L-2*S_{min}$ is obtained at $S_{min}$. $i$ denotes the index at the beginning of the DNA sequence, where $i$ starts at one at the minimum, and the maximum could be up to $L-2s-r$.
Given Eqs.~(\ref{gc}) and (\ref{hairpin}),
the proposed single-stranded loss function can be formulated as the following:
\begin{equation}
	\mathcal{L}_{BC}=\frac{1}{m}\sum_{l=1}^{m}d\left(\mathcal{G}(l),\mathcal{G}^{*}\right)^{2}+\frac{\beta}{m}\sum_{l=1}^{m}d(\mathcal{H}(l),\mathcal{H}^{*})^{2},
\end{equation}
where $d(\cdot)$ is the Euclidean distance between two items.
Accordingly, the biologically stabilized loss function of RSRL for lossless DNA storage is defined as follows:
\begin{equation}\label{bslf}
\begin{aligned}
		{\mathcal L}&=\mathcal L_{\mathrm {MASK-MSE}}+\alpha \mathcal L_{BC}
\end{aligned}\end{equation}
With the loss function shown above, the proposed RSRL can learn representations with stable structures also possessed by biomolecules.
Besides, the representations for data to store are learned by RSRL in an end-to-end manner and achieve lossless DNA storage.
More details on building and training the proposed RSRL can be found in Appendix \ref{detailed-training}.

\section{Experimental evaluation}\label{experiment}
In this section, we validate the effectiveness of the proposed RSRL by comparing it with strong baselines on real-world tasks of multi-modal data storage.
Besides, the unique properties of the proposed approach are also revealed by ablation studies.

\subsection{Compared baselines}
We compare RSRL with nine strong baseline approaches, which can be divided into two categories according to the used coding methods.
Church \cite{Church2012}, Goldman \cite{Goldman2013}, Grass \cite{Grass2015}, Blawat \cite{blawat2016forward}, DNA Fountain\cite{Erlich2017}, Yin-Yang \cite{ping2022towards}, and HL-DNA \cite{li2022hldna} are coding theory-based DNA storage methods.
DJSCC \cite{wu2023deepJoint} and DNA-QLC \cite{Zhengyan2024BMC} are learning-based DNA storage approaches.
More details of the used baselines for comparison have been illustrated in Appendix \ref{detailed-baseline}.

\subsection{Tasks of DNA storage and experimental settings}
\paragraph{DNA storage tasks} Due to the cost of DNA storage, current baselines are often experimented at KB/MB data volume levels \cite{Erlich2017, ping2022towards}. 
Following the data volume settings of previous studies, we evaluate the storage performance of all approaches using five files of diverse modalities, including images, PDFs, and text files.
For fair comparisons, all the experiments are conducted at the binary data level.
Thus, the file type has basically no effect on the performance, except in the case of lossless reading and writing.
In the main content of this paper, we report the results regarding DNA storage for PDF files.
More results showing the proposed RSRL performs similarly to the PDF storage tasks have been reported in Appendix \ref{detailed-other-muti-data}.

\paragraph{Experimental settings} To fulfill the task of multi-modal DNA storage, the proposed RSRL performs $RS(64, 48)$ to pre-coding in the $GF(2^8)$ field. 
The input dimension of the RS encoder is $M*48$. The input files are first converted to matrix form, and the output dimension is $M*64$ after being coded by RS. 
After reshaping the dimension of the file matrix to $N*32*64$ vector, it serves as input to a Transformer with two layers and four heads, which will learn representations for the subsequent DNA storage tasks. 
The learned representations are then encoded as DNA sequences according to Table \ref{binary-code}. 
Hyperparameters $\alpha$ and $\beta$ in biologically stabilized loss functions are set to 16.67 and 0.058, respectively, determined through cross-validation. 
As for the settings of compared baselines, we use the ones recommended in previous studies.
In Appendix \ref{detailed-baseline}, we provide the settings of all baselines.

\subsection{Evaluation metrics}
The performance of DNA storage mainly involves data consistency and efficiency during read/write process, and stability of the encoded DNA sequences.
In our experiments, data consistency can be directly evaluated by checking whether the learning and DNA encoding process is lossless.
Data read/write efficiency is evaluated by encoding methods, net information density, error rates, and coding speed.
As for the metrics of stability, we use minimum free energy (MFE) and melting temperature (Tm) to evaluate all approaches in our experiment.
These evaluation metrics can comprehensively reveal the performances of all approaches.
We provide the detailed definitions of these used metrics in Appendix \ref{detailed-metrics}.

\subsection{Comprehensive analysis of DNA storage performance}

\begin{table}[htbp]
\tiny
	\centering
	\caption{Summary of methods for coding information into DNA}
	\begin{tabular}{@{}p{1.5cm}p{1.5cm}p{1.5cm}p{1.5cm}p{1cm}p{1cm}p{1cm}p{1cm}p{0.75cm}p{1cm}@{}}
		\toprule
		\textbf{Method} & \textbf{Coding Method} & \textbf{Error correction strategy} & \textbf{Net information density (bits/nt)} & \textbf{GC} (\%) & \textbf{Homopolymer length (nt)} & \textbf{Avoidance of paired} & \textbf{File type} & \textbf{Lossless} \\
		\midrule
		\textbf{Church} & Direct mapping & - & 0.94 & 39–61 & 3 & × &  All Type & 1.00 \\
		\textbf{Goldman} & Ternary Huffman & Repetition & 1.48 & 39–60 & 1 & × &  All Type & 1.00 \\
		\textbf{Grass} & Galois +Rotation & RS & 1.56 & 36–62 & 3 & × &  All Type & 1.00 \\
		\textbf{Blawat} & Segment mapping & RS & 1.40 & 24–60 & 3 & × &  All Type & 1.00 \\
		\textbf{DNA Fountain} & DNA Fountain & Fountain & 1.23 & 39–62 & 4 & × & All Type & 1.00 \\
		\textbf{Yin-Yang} & Yin-yang & RS & 1.36 & 40–60 & 4 & × &  All Type & 1.00 \\
		\textbf{HL-DNA} & Quater-mapping & Barrier & 1.85 & ~51 & Have ”AA” & \checkmark &  Image & 0.896 \\
		\textbf{DJSCC} & CNN & - & - & 45-55 & \textasciitilde 5 & × & Image & 0.841 \\
		\textbf{DNA-QLC} & Conv+VAE & LC & \textbf{2.90} & 50 & 2 & × & Image & 0.926 \\
		\textbf{RSRL} & Transformer & \textbf{RS\&MSAK} & 1.75 & \textbf{\textasciitilde 50} & \textbf{\textasciitilde 3} & \textbf{\checkmark} &  \textbf{All Type} & \textbf{1.00} \\
		\bottomrule
	\end{tabular}%
	\label{table1}%
\end{table}%

We compare the overall performance of DNA storage obtained by the proposed RSRL and other advanced approaches.
The corresponding results have been listed in Table \ref{table1}.
As the table shows, RSRL demonstrates a significant advantage in net information density compared to lossless coding theory-based methods.
Compared to Goldman, RSRL achieved an 18\% improvement in Net information density.
Although learning-based approaches like DNA-QLC may obtain a higher net information density, they are not applicable for multi-modal data storage as their representation learning is not lossless.
Besides, DNA-QLC and DJSCC are computationally demanding as they stack many CNN layers for learning data representations.
The proposed RSRL is the only learning-based model that can efficiently learn lossless representations with desirable net information density.
And RSRL is the only learning-based model applicable for storing multi-modal data in DNA.

\begin{figure}[!t]
    \centering
    \begin{minipage}[t]{0.425\textwidth}
        \centering
        \includegraphics[width=\textwidth]{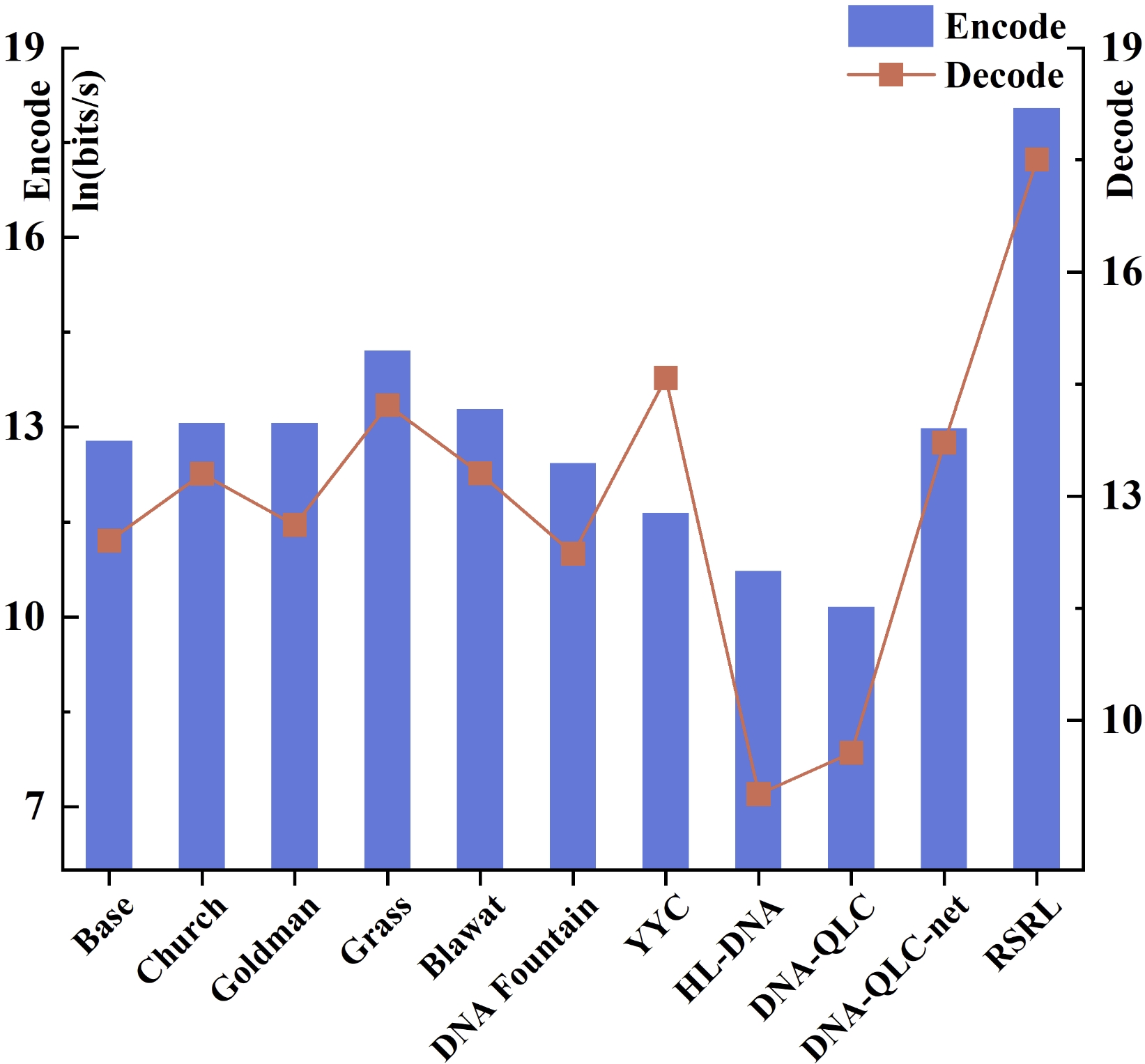}
        \caption{Comparison of Encoding Speed between RSRL and other baselines.}
        \label{fig_4}
    \end{minipage}
    \hfill
    \begin{minipage}[t]{0.425\textwidth}
        \centering
        \includegraphics[width=\textwidth]{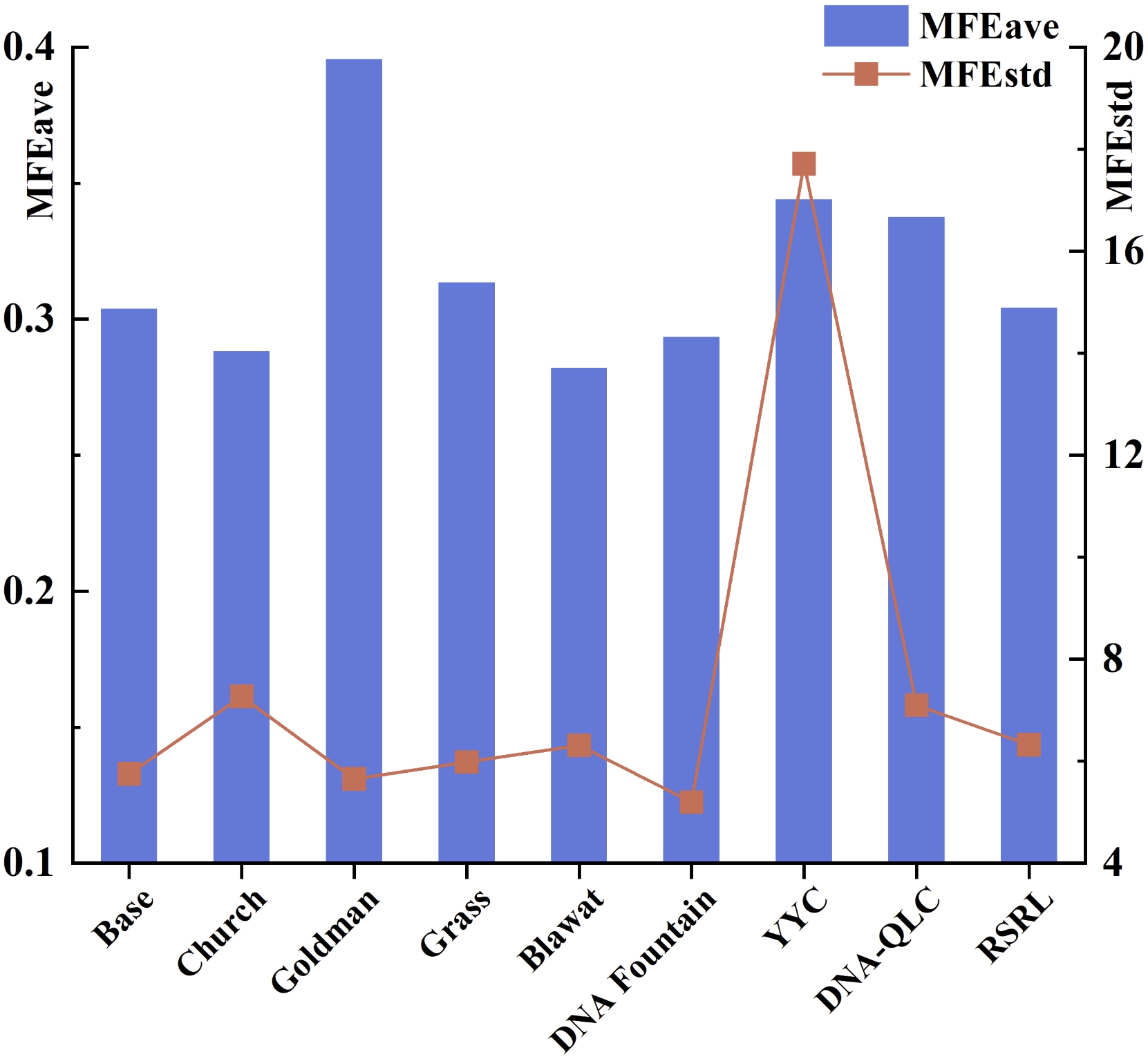}
        \caption{Comparison of mean and standard deviation of MFE between RSRL and other baselines.
        }
        \label{fig_5}
    \end{minipage}
\end{figure}
Regarding data loss, both DJSCC and DNA-QLC use convolution to compress the input image. It is known that this processing introduces loss, which we quantify using the Structural Similarity Index (SSIM). In the best-case scenario, the SSIM values for images stored by DJSCC and DNA-QLC are 0.841 and 0.926, respectively, both falling short of 1. In contrast, the proposed RSRL achieves lossless storage of multimodal data, as shown in Table \ref{table1}.
Obtaining such results is mainly because RSRL is the first learning-based approach to DNA storage incorporated with Reed-Solomon (RS) coding as an error correction strategy.
Moreover, we further propose the MASK-MSE loss based on RS coding, which converts random errors that are difficult for RS to handle into burst errors (Fig.~\ref{fig_2}B), thereby maximizing the error correction potential of RS codes.

From Table \ref{table1}, we also observe that the proposed RSRL can make a good balance between the performance of DNA storage and the indicators of biological stability.
This is because RSRL additionally adopts the proposed single-stranded loss functions based on structural biology, achieving constraint satisfaction through a learning approach and overcoming the limitations of learning-based approaches (e.g., DJSCC) in terms of the number and accuracy of constraints.

Encoding speed directly impacts read-write latency in DNA storage. 
The comparisons of encoding speed between RSRL and other baselines are depicted in Fig.~\ref{fig_4}, where the proposed RSRL demonstrates the highest speed of encoding data for DNA storage.
RSRL can encode more data per unit of time than other baselines because it adopts a lightweight network structure (See Appendix \ref{detailed-setting}). 
In our experiments, we additionally design DNA-QLC-net, which is a variant of DNA-QLC and only records the time cost by the neural network.
As the figure shows, DNA-QLC-net is still much slower than the proposed RSRL due to its complex network structure.

\begin{figure}[!t]
    \centering
    \begin{minipage}[t]{0.425\textwidth}
        \centering
        \includegraphics[width=\textwidth]{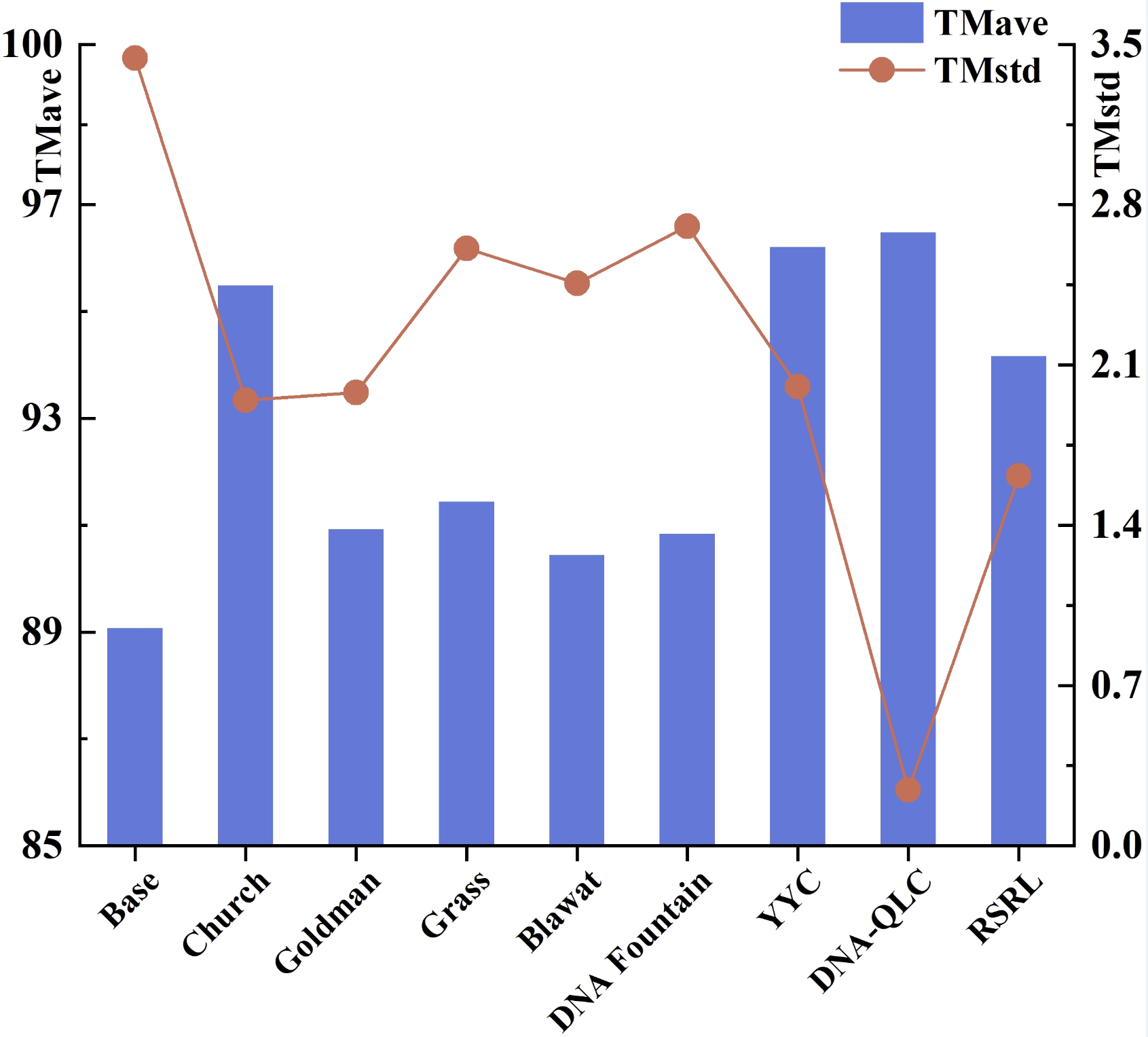}
        \caption{Comparison of mean and standard deviation of Tm between RSRL and other baselines.
        The left vertical coordinate is the value of Tmave, and the right is that of Tmstd.}
        \label{fig_6}
    \end{minipage}
    \hfill
    \begin{minipage}[t]{0.425\textwidth}
        \centering
        \includegraphics[width=\textwidth]{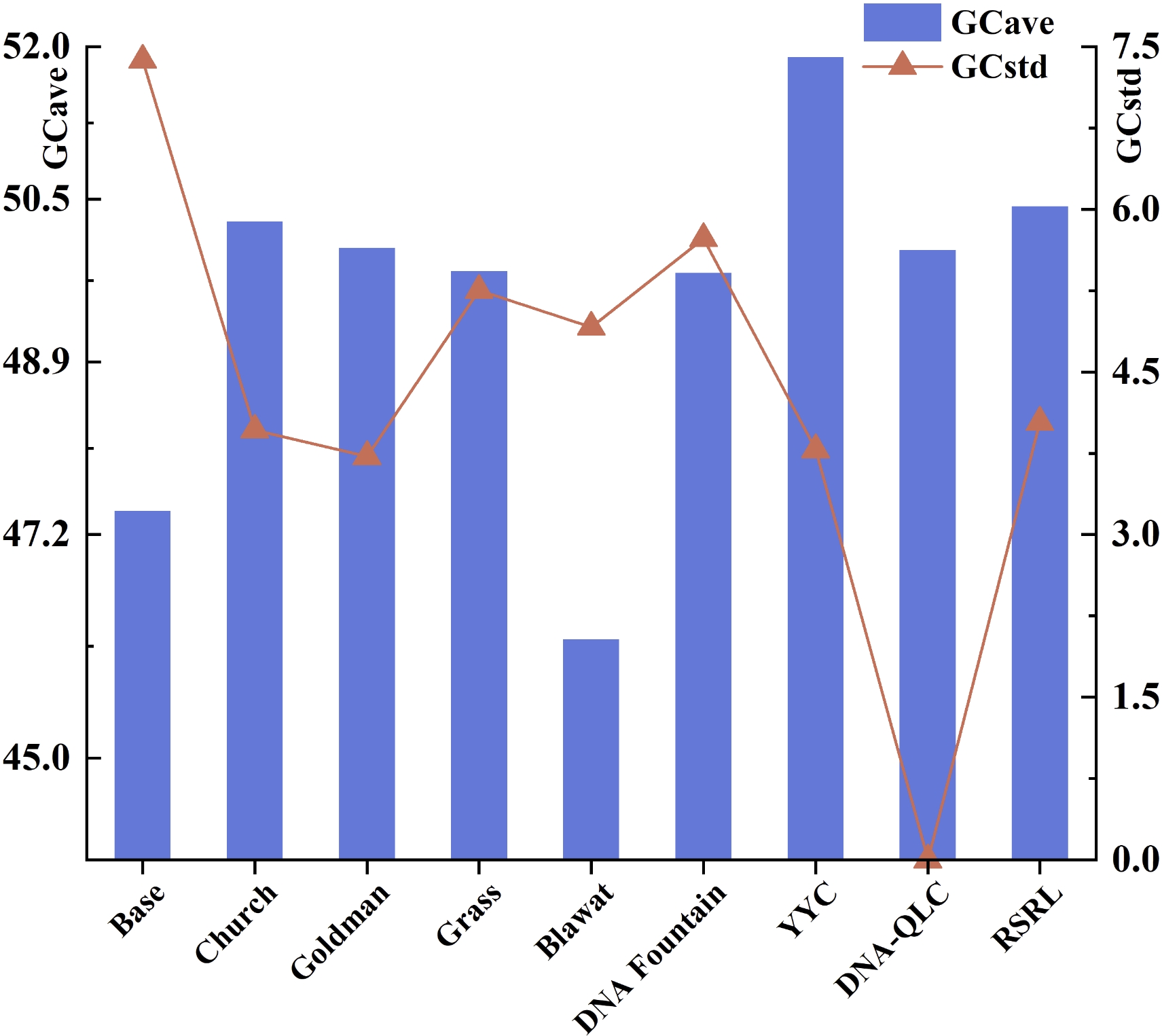}
        \caption{Comparison of mean and standard deviation of GC content between RSRL and other baselines.
        The left vertical coordinate is the value of GCave, and the right is that of GCstd.}
        \label{fig_7}
    \end{minipage}
\end{figure}

\begin{figure}[!t]
	\centering
	\includegraphics[width=0.75\textwidth]{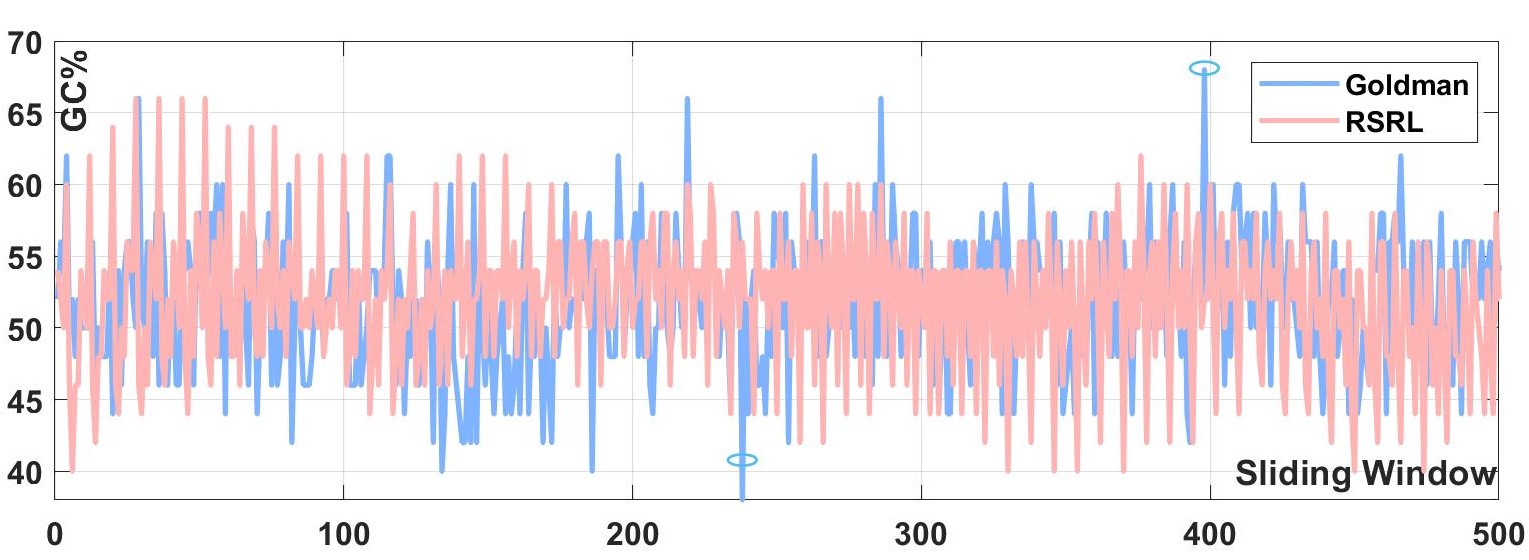}
    \vspace{1.0em}
	\caption{Distribution of local GC content under a sliding window.}
	\label{fig_8}
\end{figure}

\subsection{Thermodynamic comparisons of coding}

Thermodynamic changes can better reflect the essence of biochemical reactions, consistently interweaving with biochemical reactions, thus more directly manifesting the stability and performance of DNA sequences.
In DNA storage tasks, DNA sequences can be evaluated based on thermodynamic properties such as free energy, melting temperature, and GC content. 
In our experiments, we compare RSRL with other baseline methods in terms of minimum Gibbs free energy, melting temperature, GC content, and local GC content. The results are presented in Figs.~\ref{fig_5}-\ref{fig_8}.

\paragraph{Minimum free energy} Energy changes directly reflect the intrinsic variations in biochemical reactions, thus indicating the stability of DNA sequences for data storage.
In our experiment, we use Gibbs standard free energy ($\Delta G$), a widely accepted indicator to reflect energy changes \cite{Wang2018magicMFE}. 
In Fig. \ref{fig_5}, the corresponding results of minimum Gibbs free energy are depicted.
As shown in Fig.~\ref{fig_5}, RSRL exhibits a smaller MFEave compared to Goldman, Grass, Yin-Yang, and DNA-QLC.
A lower MFE indicates that the DNA sequences encoded by RSRL are more stable. 
While MFEave of RSRL is similar to Church and Blawat, its MFEstd is more advantageous, indicating that the quality of the DNA sequences encoded by RSRL tend to be stable, with less influence from outliers. 
Particularly compared to DNA-QLC (the learning-based method), both MFEave and MFEstd obtained by RSRL are lower by over 11\%, signifying the superior performance of the network model and biologically stabilized properties adopted by RSRL.

\paragraph{Melting temperature and GC content} Throughout the entire process of DNA storage, factors such as DNA assembly, PCR amplification, and storage stability necessitate that the melting temperature (Tm) 
\cite{li2020constraining} and GC content of oligonucleotides exhibit minimal deviation. 
Therefore, we compare the Tm and GC content of RSRL with other baseline methods. 
When Tm fell within the range of (85-95), particular attention is given to the standard deviation of Tm (Tmstd). Smaller TMstds ensure smooth data reading and writing processes in DNA storage.
In Fig.~\ref{fig_6}, the Tmstd of RSRL is shown to be significantly lower than that of other coding theory-based schemes. 
The Tmstd of RSRL is higher than that of DNA-QLC, which may result from DNA-QLC's time-consuming hard coding after completing the learning of representations, limiting the GC content to 50\%, as shown in Fig.~\ref{fig_7}.
However, directly setting GC content as 50\% does not give rise to a significant boost of performance, compared to 48-50\%, which has a negligible effect in DNA storage \cite{yang2020gc,ping2022towards}.
So, the optimal value of GC content is generally considered as around 50\%.
In Fig.~\ref{fig_7}, we compare the average (GCave) and variance (GCstd) of GC content, showing that all methods maintain GC content within the range of 46-52\%, essentially meeting the GC content constraint and validating the effectiveness of the encoding strategy. 

The GC content is calculated by considering the entire DNA sequence as the smallest unit in Fig.~\ref{fig_7}, which might overlook the impact of local GC content \cite{cao2022adaptive}. 
Therefore, we additionally analyze local GC content, comparing RSRL with the Goldman method, which is most similar to RSRL in terms of global GCave and GCstd. 
From Fig.~\ref{fig_8}, it is evident that the local GC content of the proposed RSRL is smoother. While local GC content of Goldman have more outliers, potentially resulting in lower off-machine quality in Illumina sequencing data, and affecting the consistency of DNA storage data reading and writing.

\begin{table}[t]
    \tiny
	\centering
	\caption{Ablation study of RSRL}
	\begin{tabular}{ccccccccccc}
		\toprule
		\textbf{Method} & \textbf{Reconstruction Rate} & \textbf{Block Failure Rate} & \textbf{MFEave} & \textbf{MFEstd} & \textbf{Tmave} & \textbf{Tmstd} & \textbf{GCave} & \textbf{GCstd} \\
		\midrule
		\textbf{RSRL-No MASK} & 93.42\% & 9.6\% & -18.72 & 4.32 & 85.86 & 2.20 & 38.22 & 3.92 \\
		\textbf{RSRL-No GC\&pair} & \textbf{100}\% & 0 & -24.7 & 4.24 & 91.03 & 2.58 & 48.33 & 5.76 \\
		\textbf{RSRL-No GC\&pair\&MASK} & 98.76\% & 2.4\% & -20.66 & \textbf{3.66} & 88.90 & 1.74 & 44 & 3.84 \\
        \textbf{RSRL} & \textbf{100}\% & \textbf{0} & \textbf{-34.37} & 6.31 & \textbf{94.16} & \textbf{1.61} & \textbf{50.42} & \textbf{4.03} \\
    \bottomrule
	\end{tabular}%
	\label{table2}%
        \vspace{-1.0em}
\end{table}%

\subsection{Ablation study}

In this subsection, we conduct ablation studies to show the effect of each module in the proposed RSRL on DNA storage tasks.
Specifically, we systematically analyze the effects of MASK-MSE and biologically stabilized loss functions on the performance of RSRL.
The results have been listed in Table \ref{table2}.
We first compare RSRL without MASK-MSE (RSRL-No-MASK) and RSRL. 
Results indicate that RSRL-No-MASK fails to achieve lossless data reconstruction, losing approximately 9.6\% of data blocks. 
Also, there is an evident performance gap regarding MFEave and GCave when comparing RSRL-No-MASK with RSRL.
We then evaluate the performance of RSRL without biologically stabilized functions (RSRL-No GC\&pair). 
Despite the combined effect of Mask and RS codes, this version of RSRL successfully recovered data but exhibits a noticeable decline in thermodynamic results. 
Compared to RSRL, the MFEave of RSRL-No GC\&pair is increased by almost 50\%, indicating insufficient stability in DNA sequence double-strand binding, which may lead to errors during DNA storage. 
At last, we evaluate the performance of the variant only adopting the conventional MSE loss (RSRL-No GC\&pair\&mask).
Although it surpasses RSRL-No MASK in terms of reconstruction rate, it still fails to reconstruct the data completely. Due to the use of conventional loss functions, the learning process of RSRL-No GC\&pair\&mask lacks focus and poses risks to read-write consistency.

\section{Conclusion}
In this paper, we have proposed Reed-Solomon coded single-stranded representation learning (RSRL), a novel end-to-end model for learning represnetations for DNA storage.
Unlike existing learning-based approaches to DNA storage, RSRL incorporates an error-correction codec and stable biological structures into the process of learning representations for data storage.
Representations learned by RSRL possess remarkable structural properties like biomolecules in biont and are, therefore, highly durable, dense, and lossless for subsequent storage tasks.
The proposed RSRL has been compared with both coding theory and learning-based methods for DNA storage. 
The obtained experimental results demonstrate that RSRL can outperform prevalent approaches in the tasks of representation learning for multi-modal data.
In the future, we will further improve the proposed RSRL by identifying more efficient strategies to incorporate error-correction codes into neural networks and formulating more efficient biologically informed loss functions for model training.

\bibliographystyle{plain}
\bibliography{ref}

\newpage
\appendix

\section{More details on DNA data storage}\label{intro-DNAs}

The International Data Corporation (IDC) predicts that the capacity of the global data circle will increase to 175 zettabytes by 2025. Current storage systems are faced with high cost and huge energy consumption.
In contrast, DNA is a highly parallel, low-cost storage medium with great storage potential. 
Different from traditional storing medium that are replaced every few years, DNA is very stable in decades or centuries, and can be easily replicated and stored based on methods of molecular biology.

There are five steps in DNA data storage, including encoding information into DNA codewords (encoding), synthesizing DNA from the sequence (writing), storage, DNA sequencing (reading), and decoding.
For encoding, fountain, rotation, and Huffman code are commonly used methods.
Methods that are based on modern AI techniques have also been proposed for encoding purposes.
Popular techniques for DNA synthesis can then be used to synthesize the sequences storing real-world data in DNA.
After that, DNA molecules of real-world data can be stored in test tubes or in the form of dry powder for a very long period.
To read the data from DNA, prevalent sequencing methods, e.g., Illumina sequencing and nanopore sequencing can be used.
After processing the sequenced DNA with clustering and assembly techniques, the original data can finally be recovered by decoding, which is completed inversely by the previously used encoding method.

Currently, the main bottlenecks existing in DNA storage are cost and read/write latency.
DNA storage is costly due to DNA synthesis and sequencing.
Issues of read/write latency in DNA storage are mainly due to codec and corresponding biotechnologies.
It is seen that efficient and robust codec algorithms can reduce not only the cost by improving the encoding rate but also the read-write latency by reducing errors in DNA storage.
In this paper, we propose a novel learning-based model that can significantly reduce the read and write delay from the codec stage.
Besides, the proposed model provides an economical solution to DNA storage by reducing the error rate and improving the encoding rate.

\section{More details on the baselines} \label{detailed-baseline}
Based on the coding method, the baselines can generally divided into two categories, i.e., coding theory based on learning based approaches.
Methods based on coding theory have the advantage of predictable results and complete proof of theory.
\begin{itemize}
	\item Church \cite{Church2012}: This method proposes encoding one binary bit per base (A or C for 0, G or T for 1) to encode bitstreams directly into DNA sequences.
	\item Goldman \cite{Goldman2013}: This approach uses the Huffman trinomial tree to analyze binary files to be transcoded based on the frequency of occurrence of individual bytes. The binary sequences (0/1) are converted to the corresponding ternary sequences (0/1/2), which are subsequently mapped to the corresponding DNA sequences according to the ternary mapping model.
	\item Grass \cite{Grass2015}: This approach combines the Galois field with the DNA codon wheel style base mapping rules to propose a coding algorithm that avoids the length of a single base being greater than three.
	\item Blawat \cite{blawat2016forward}: This method uses the byte as the basic unit of base conversion and maps eight bits of information into five nucleotides. The first six bits are fixed conversion portions mapped to nucleotides A, C, G, and T. The last two bits are optional conversion portions. This design limits the maximum length of the homopolymer to three.
	\item DNA Fountain \cite{Erlich2017}: This approach preprocesses binary information into a series of non-overlapping fragments, randomly selects a variable number of sequence fragments for heterogeneous operation based on Luby Transform, and appends a fixed-length seed to form a droplet.
	\item Yin-Yang \cite{ping2022towards} : This method provides a dynamic combinatorial coding scheme that combines two independent coding rules (called "yin" and "yang") into a single binary sequence, thus compressing two bits into a single nucleotide.
	\item HL-DNA \cite{li2022hldna}: This approach proposes a hybrid lossy and lossless image storage scheme implemented by quaternary mapping. HL-DNA uses about 300 nt as the length of DNA strands and four extra nucleotides, and it is a hybrid lossy/lossless encoding scheme.
\end{itemize}
Recently, attention has been paid to learning-based DNA storage. However, due to the information loss inherent in neural network learning, current learning-based DNA storage schemes are lossy and can only be used for storing multimedia data such as images.
\begin{itemize}
	\item DJSCC \cite{wu2023deepJoint}: 
    This approaches uses a convolutional neural network for the DNA encoding and decoding process and reconstructs the loss function by optimizing for GC content and homopolymer constraints. DJSCC used input size $(32,32,3)$, $R(nt/pixel)$ is 1/8,  1) Encoder: $CV (32, 1, 3)$, $BN$, $ReLU$; $CV (64, 32, 3)$, $BN$, $ReLU$; $CV (63, 128, 3)$; $FC (32). 2)$ Decoder: $FC (1152)$, $DC (128, 64)$, $BN$, $ReLU$; $DC (64, 32)$, $BN$, $ReLU$, $DC (32, 1)$, Sigmoid.
	\item DNA-QLC \cite{Zhengyan2024BMC}: This approach uses the quantized ResNet VAE (QRes-VAE) model and LC for image compression. It differs from VAE-QC in that it also designed an error correction module to improve the storage system's robustness. DNA-QLC used the QRes-VAE, and based on ResNet VAE, all parameters are consistent with the original text.
\end{itemize}

As for the settings of RSRL, the input dimension, hidden layer, and compression dimension are set as (64,32,56), respectively, with four attention headers, two layers of encoders, and the Adam optimizer.
The experimental environment for this work is RTX3090, 256G RAM, i7 9700k. 

\section{More analysis and discussions on the performance of DNA storage}\label{overview-performancene}
Here, we further analyze the results in Table \ref{table1}. 
Since higher biological stability of DNA sequences can reduce the probability of errors occurring during the storage process, this section mainly analyzes the GC content, homopolymers, and hairpin structure of the encoding results.

\paragraph{GC content and homopolymers} Most storage schemes can meet the GC content constraint. 
However, the fluctuations (large deviations) of GC content obtained by Grass and Blawat are evident.
Previous studies have shown that deviations of GC content have a significant effect on the melting temperature \cite{ross2013characterizing}. Thus, the PCR yield and sequencing accuracy of these two approaches are reduced.
In contrast, RSRL, DNA-QLC, and HL-DNA perform robustly when evaluated by GC content (~50\%), demonstrating that these approaches may cause fewer errors during DNA sequencing.
Homopolymers during synthesis can cause difficulties, while during sequencing, they may result in gaps such as $AAAA$ being misread as $AAA$, affecting data consistency. However, overly strict homopolymers could impact base utilization since the total number of base combinations is fixed. RSRL limited homopolymers to around three, achieving a good balance between base utilization and data consistency in reading and writing.

\paragraph{Hairpin structure} A hairpin structure is a distinctive secondary structure in DNA, 
where two base pairs are held together by hydrogen bonds \cite{li2020constraining}. 
The hairpin structure can increase the error rate when reading and replicating in DNA storage, thus influencing the performance.
It can be seen from Table \ref{table1} that RSRL is one of the few approaches to DNA storage, considering the impact of hairpin structures.
Although DJSCC can satisfy GC content and homopolymer constraints through learning, its deviations of GC content and homopolymer limitations are higher than those of RSRL.
DNA-QLC attempts to force the learned representations to satisfy the biological constraints by post-processing.
Thus, it is not an end-to-end learning approach.
Moreover, existing learning approaches like DJSCC and DNA-QLC, can only store multimedia data, e.g., image data, due to the information loss during model training.
Compared with existing learning-based approaches, the proposed RSRL is the first end-to-end model that considers stable properties of biological structures.
Thus, RSRL performs much better than existing learning-based methods regarding diverse metrics of DNA storage.

\section{Additional results of multimodal data encoding}\label{detailed-other-muti-data}

Although any data type is binary at input to the model, RSRL has no bias to the data type, that is, the data type has minimal influence on the encoding result. However, in order to further illustrate the applicability of RSRL to multi-modal data, we also verified this point through experiments, as shown in Tables~\ref{table-otherfile-PDF}-\ref{table-otherfile-Fig}. The evaluation metrics are the same as the main text, and the results show that RSRL is unbiased in terms of data types. In particular, the unit of encoding and decoding time is bits/s.

\begin{table}[t]
    \tiny
	\centering
	\caption{Results of PDF data encoding.}
	\begin{tabular}{ccccccccccc}
		\toprule
		\textbf{Method} & \textbf{NID}& \textbf{MFEave} & \textbf{MFEstd} & \textbf{Tmave} & \textbf{Tmstd} & \textbf{GCave} & \textbf{GCstd} &\textbf{Encoding time} & \textbf{Decoding time}\\
		\midrule
		\textbf{Grass}  & 1.56 & -31.25 & 6.37 & 91.55 & 2.78 & 49.65 & 5.87 & 1.32E6 & 2.00E6 \\
		\textbf{Blawat}  & 1.40 & -23.39 & 6.58 & 90.65 & 1.61 & 46.47 & 5.42 & 6.89E7 & 4.00E7 \\
            \textbf{DNA Fountain} & 1.23 & -23.31 & 5.32 & 90.55 & 2.77 & 49.82 & 5.83 & 1.56E5 & 1.83E5\\
		\textbf{DNA-QLC} & - & - & - & - & - & - & - & - & - \\	\textbf{RSRL} & 1.75 & -34.37 & 6.31 & 94.16 & 1.61 & 50.42 & 4.03 & 6.89E7 & 4.00E7 \\
    \bottomrule
	\end{tabular}%
	\label{table-otherfile-PDF}%
\end{table}%

\begin{table}[t]
    \tiny
	\centering
	\caption{Results of text data encoding.}
	\begin{tabular}{ccccccccccc}
		\toprule
		\textbf{Method} & \textbf{NID}& \textbf{MFEave} & \textbf{MFEstd} & \textbf{Tmave} & \textbf{Tmstd} & \textbf{GCave} & \textbf{GCstd} &\textbf{Encoding time} & \textbf{Decoding time}\\
		\midrule
		\textbf{Grass}  & 1.56 & -25.70 & 5.98 & 91.43 & 2.60 & 49.79 & 5.25 & 1.48E6 & 1.49E6\\
  		\textbf{Blawat}  & 1.40 & -38.77 & 7.4 & 96.57 & 1.74 & 58.15 & 3.59 & 5.86E5 & 6.0E5\\
    	\textbf{DNA Fountain} & 1.23 & -26.64 & 5.98 & 92.19 & 2.54 & 49.57 & 5.44 & 8.49E3 & 3.45E4\\
		\textbf{DNA-QLC} & - & - & - & - & - & - & - & - & - \\
		\textbf{RSRL} & 1.75 & -32.07 & 5.67 & 94.54 & 2.36 & 51.28 & 6.47 & 6.32E7 & 4.11E7\\
    \bottomrule
	\end{tabular}%
	\label{table-otherfile-txt}%
\end{table}%

\begin{table}[t]
    \tiny
	\centering
	\caption{Results on image data encoding.}
	\begin{tabular}{ccccccccccc}
		\toprule
		\textbf{Method} & \textbf{NID}& \textbf{MFEave} & \textbf{MFEstd} & \textbf{Tmave} & \textbf{Tmstd} & \textbf{GCave} & \textbf{GCstd} &\textbf{Encoding time} & \textbf{Decoding time}\\
		\midrule
		\textbf{Grass}  & 1.56& -29.09 & 5.42 & 92.86 & 1.93 & 51.37 & 3.73 & 1.43E6 & 1.42E6\\
		\textbf{Blawat}  & 1.40 & -24.25 & 6.85 & 90.43 & 2.45 & 45.35 & 5.77 & 1.22E6 & 1.69E6\\
    	\textbf{DNA Fountain} & 1.23& -34.37 & 6.31 & 94.16 & 1.61 & 46.42 & 4.29 & 6.10E5 & 4.03E7\\
		\textbf{DNA-QLC} & 2.90& -69.19 & 7.09 & 96.47 & 0.24 & 50 & 0 & 2.58E4 & 1.43E4 \\		
            \textbf{RSRL} & 1.75 &  -33.92 & 4.06 & 94.97 & 2.20 & 51.57 & 4.89 & 6.30E7 & 3.93E7\\
    \bottomrule
	\end{tabular}%
	\label{table-otherfile-Fig}%
\end{table}%

\section{Details on the metrics of DNA thermodynamics
}\label{detailed-metrics}
\subsection{Minimum free energy}

The Gibbs Free Energy contains two important thermodynamic parameters, entropy change and enthalpy change: $\Delta G = \Delta H - T\Delta S
$, where $T$ is the temperature.
The minimum Gibbs free energy is the minimum value of the standard free energy of all possible secondary structures in the DNA sequence.
Secondary structures with lower Gibbs free energy are more stable. 
Therefore, the minimum free energy (MFE) could be used to assess the quality of DNA sequences. 
Let $\Delta G(s, s^{\prime})$ denote the Gibbs free energy value of DNA sequence $s$, where $s^{\prime}$ is its complementary strand, which could be calculated using the PairFold method \cite{santalucia1998DNApairfold}:
\begin{equation}\label{mfe}
MFE=\min\{\Delta G(u,v),\Delta G(u,v^{\prime}),\Delta G(u^{\prime},v^{\prime})\},
\end{equation}
where $\Delta G(u,v)$, $\Delta G(u,v^{\prime})$, and $\Delta G(u^{\prime},v^{\prime})$ are the Gibbs free energy between $u$ and $v$, and that between their complements, respectively.
Given the the number of sequences and their lengths, the average MFE (MFEave) can be computed as the following: 
\begin{equation}
MEF_{ave}=\frac1n\sum_{i=1}^n\frac{MFE_i}{L_i},
\end{equation}
where $n$ is the number of DNA sequences, and $L_i$ is the length of DNA sequence $i$. Accordingly, we are able to obtain the standard deviation of MFE of each approach (MFEstd).

\subsection{Melting temperature}

In this paper, the melting temperature (Tm) is calculated as the following:

\begin{equation}\mathrm{Tm}=\Delta H^\circ/(\Delta S^\circ+R\ln C_\mathbf{T})-273.15,\end{equation}

where $R$ is the gas constant (1.987 cal/K·mol), $C_\mathbf{T}$ is the total oligonucleotide strand concentration, $\Delta H^\circ$ and $\Delta S^\circ$ can be obtained by search the Unified oligonucleotide table \cite{santalucia1998unified}. 
The Tm represented the temperature required for a DNA sequence to transition from a double-stranded structure to a single-stranded structure. 
When Tm falls within the range of (85-95), we should pay more attention to the standard deviation of Tm (Tmstd), as stable Tm values indicate orderly progression of assembly and PCR processes.
Then, A smoother data reading and writing process in DNA storage can be achieved.

\section{Network structure of RSRL and other learning-based approaches}\label{detailed-setting}

\begin{table*}[htbp]
	\centering
	\caption{Comparison of the number and complexity of model parameters.}
        \vspace{1.0em}
	\begin{tabular}{@{}cccccc@{}}
		\toprule
		Model & Input size & Total params & Total memory (MB) & Total Madd(M) & Total Flops(M) \\
		\midrule
		DJSCC & (32,32,3) & 54921 & 0.24 & 7.47M & 1.76M \\
		DNA-QLC & (64,64,1) & 3538944 & 13.5 & 906M & 1812M \\
		RSRL & (32,32,64) & 49152 & 0.1875 & 0.049M & 0.098M \\
		\bottomrule
	\end{tabular}%
	\label{table6}%
\end{table*}%

In this section, we compare the network structure of the proposed RSRL with other learning-based approaches, including DJSCC \cite{wu2023deepJoint} and DNA-QLC \cite{Zhengyan2024BMC}.
Regarding computational complexity, we mainly consider Total Madd and Total Flops, which represent the number of multiply-accumulate and floating-point operations.
Regarding space complexity, we mainly consider the total memory used by each model (Total Memory) and the total number of parameters of each model (Total Params).
The comparative results have been listed in Table \ref{table6}.
As the table shows, the proposed RSRL achieves much lower than any other two approaches, demonstrating it is very efficient in learning representations for DNA storage.
Together with the performance results of DNA storage, we conclude that a lightweight learning model with proper biological inspirations (e.g., the biologically stabilized loss proposed in this paper) can still perform robustly in diverse DNA storage tasks.

\begin{table}[htbp]
	\centering
	\caption{Binary streams coded as base sequences}
	\begin{tabular}{ccccc}
		\hline
		\textbf{} & \textbf{00} & \textbf{01} & \textbf{10} & \textbf{11} \\ \hline
		\textbf{00} & AT & AG & AC & AA \\ 
		\textbf{01} & TA & TC & TG & TT \\ 
		\textbf{10} & GG & GA & GT & GC \\ 
		\textbf{11} & CC & CT & CA & CG \\ \hline
	\end{tabular}\label{binary-code}
\end{table}

\section{End-to-end training}\label{detailed-training}
In this section, we provide more details on building the proposed RSRL. In the encoding phase, a file is first processed into a binary data stream. 
Since the domain of RS codes is characters, the binary data stream is grouped and converted into hexadecimal characters. RS codes are performed on the hexadecimal characters in the Galois domain of $2^8$. The encoding result is then reshaped as 32*32*64 as an input to the Transformer, which has a 2-layer encoder with four multi-head attention. 
The Transformer can learn a 32*32*56 low-dimensional representation for each file, which is normalized to 0 or 1 by a liner layer.
The last 8 bits of the representation (an 8-bit binary could be transcoded into 1-bit hexadecimal characters) are masked for the subsequent error correction.
Then, the normalized representation is transcoded to a DNA sequence by Table \ref{binary-code}.

For the decoder, the input is a DNA sequence, inverted to a binary matrix by Table \ref{binary-code}. 
The structure of the decoder is similar to that of the encoder, except that a masking mechanism and encoder-decoder attention are used to make the decoder pay more attention to the different parts of the input sequence. The raw tensor vector output from the Transformer decoder is converted into hexadecimal characters and fed into the RS decoder for decoding. 
Due to the Mask operation at the time of encoding, the errors are concentrated in the last 8 bits of the tensor so that RS codes can correct all errors in the last 8 bits.
Finally, the RS decoded hexadecimal characters are converted to a binary stream to complete the reconstruction of the original file.



\end{document}